%% file: 00_main.tex
\documentclass[10pt, a4paper]{article}
\usepackage{lrec2022} % this is the new LREC2022 Style
\usepackage{multibib}
\newcites{languageresource}{Language Resources}
\usepackage{graphicx}
\usepackage{tabularx}
\usepackage{soul}
\usepackage{titlesec}
\titleformat{\section}{\normalfont\large\bfseries\center}{\thesection.}{1em}{}
\titleformat{\subsection}{\normalfont\SmallTitleFont\bfseries\raggedright}{\thesubsection.}{1em}{}
\titleformat{\subsubsection}{\normalfont\normalsize\bfseries\raggedright}{\thesubsubsection.}{1em}{}
\renewcommand\thesection{\arabic{section}}
\renewcommand\thesubsection{\thesection.\arabic{subsection}}
\renewcommand\thesubsubsection{\thesubsection.\arabic{subsubsection}}
\usepackage{epstopdf}
\usepackage[utf8]{inputenc}

\usepackage{hyperref}
\usepackage{xstring}

\usepackage{color}

\usepackage[fleqn]{amsmath}
\usepackage{mathtools, amsthm, amssymb, amsfonts}
\usepackage{xfrac}
\usepackage{booktabs, dcolumn}
\usepackage{colortbl}
\usepackage{enumitem}
\usepackage{comment}

\usepackage[normalem]{ulem}

\usepackage{xcolor}

\definecolor[named]{Chemical}{rgb}{0.3,0.7,0.3}
\definecolor[named]{Disease}{rgb}{0.95,0.38,0.21}
\definecolor[named]{Gene}{rgb}{0.4,0.23,0.72}

\definecolor[named]{BrickRed}{rgb}{0.71,0.2,0.11}
\definecolor[named]{NoteBlue}{rgb}{0.13,0.51,1.0}
\definecolor[named]{ForestGreen}{rgb}{0.0,0.7,0.2}
\newcount\Comments  % 0 suppresses notes to selves in text
\definecolor{darkgreen}{rgb}{0,0.5,0}
\definecolor{darkred}{rgb}{0.7,0,0}
\definecolor{teal}{rgb}{0.3,0.8,0.8}
\definecolor{blue}{rgb}{0,0,1}
\definecolor{purple}{rgb}{0.5,0,1}
\newcommand{\kibitz}[2]{\ifnum\Comments=1\textcolor{#1}{#2}\fi}

\newcommand*{\tsub}[1]{\textsubscript{\textnormal{#1}}}
\newcommand*{\tsup}[1]{\textsuperscript{\textnormal{#1}}}

\newcommand{\itemi}[1]{\item{\itshape #1}}

\title{A Distant Supervision Corpus for Extracting Biomedical Relationships Between Chemicals, Diseases and Genes}

\name{Dongxu Zhang\tsup{*,1}\thanks{\hspace{-4pt}{*} Equal contribution}, Sunil Mohan\tsup{*,2}, Michaela Torkar\tsup{2}, Andrew McCallum\tsup{1}}

\address{\tsup{1} University of Massachusetts, Amherst, Massachusetts, USA\\
	\tsup{2} Chan Zuckerberg Initiative, Redwood City, California, USA \\
    \{dongxuzhang, mccallum\}@cs.umass.edu\\
    smohan@chanzuckerberg.com, michaela.torkar@contractor.chanzuckerberg.com\\}

\abstract{
We introduce {\em ChemDisGene}, a new dataset for training and evaluating multi-class multi-label document-level biomedical relation extraction models.  Our dataset contains 80k biomedical research abstracts labeled with mentions of chemicals, diseases, and genes, portions of which human experts labeled with 18 types of biomedical relationships between these entities (intended for evaluation), and the remainder of which (intended for training) has been distantly labeled via the CTD database with approximately 78\% accuracy.
In comparison to similar preexisting datasets, ours is both substantially larger and cleaner; it also includes annotations linking mentions to their entities.
We also provide three baseline deep neural network relation extraction models trained and evaluated on our new dataset.
\\ \newline  }

\begin{document}

\maketitleabstract

%
% Put sections into separate files
%

\input{01_intro}
\input{03_methodology}
\input{04_corpus}
\input{05_re_task}
% Moved 'Related Work' to the end
\input{02_related_work}
\input{06_conclusion}

% ----------------------------------------------------------------------------------------------------------
\section*{Acknowledgments}

%This work is based upon work supported in part by the Center for Data Science and the Center for Intelligent Information Retrieval, and in part by the National Science Foundation under Grants No. 1763618, and in part by the National Science Foundation under Grant No. 1514053, and in part by the Chan Zuckerberg Initiative under the project “Scientific Knowledge Base Construction. Some of the work reported here was performed using high performance computing equipment obtained under a grant from the Collaborative R\&D Fund managed by the Massachusetts Technology Collaborative. Any opinions, findings and conclusions or recommendations expressed in this material are those of the authors and do not necessarily reflect those of the sponsor.

We thank the CZI Curation team for their work on the curated corpus, and members of UMass IESL and NLP groups for helpful discussion and feedback. 
This work is based upon work supported in part by the Center for Data Science and the Center for Intelligent Information Retrieval, and in part by the National Science Foundation under Grant Nos. 1763618 and 1514053, and in part by the Chan Zuckerberg Initiative under the project ``Scientific Knowledge Base Construction''. Some of the work reported here was performed using high performance computing equipment obtained under a grant from the Collaborative R\&D Fund managed by the Massachusetts Technology Collaborative. 
Any opinions, findings, conclusions or recommendations expressed in this material are those of the authors and do not necessarily reflect those of the sponsor.

% ----------------------------------------------------------------------------------------------------------

% \nocite{*}
\section{Bibliographical References}\label{reference}
%\label{main:ref}

\bibliographystyle{lrec2022-bib}
\bibliography{biblio}

\section{Language Resource References}
\label{lr:ref}
\bibliographystylelanguageresource{lrec2022-bib}
\bibliographylanguageresource{languageresource}

\end{document}

%% file: 01_intro.tex
\section{Introduction}
\label{sec:intro}

Biomedical researchers have used systems of experimentally confirmed interactions between chemicals, diseases, genes/proteins and other entities, for understanding disease mechanisms for diagnosis, e.g. \newcite{Lee-etal:2019}, for drug repurposing \cite{Gysi-etal:2021}, and even for understanding the health hazards associated with spaceflight \cite{Nelson-etal:2021}. These knowledge graphs (KGs) are often built by integrating manually curated databases like CTD\footnote{CTD: \href{http://ctdbase.org}{Comparative Toxicogenomics Database}} and DrugBank\footnote{\url{https://go.drugbank.com}}, who use domain experts to extract observed interactions from research publications and other sources. While the information in these databases is high in {\em precision}, with the growing publication rate their {\em recall} is low \cite{Baumgartner-etal:2007}. To improve coverage, researchers have resorted to automated mining of biomedical interactions from research texts, to supplement their KG \cite{Himmelstein-etal:2017}, or even to build the entire KG, e.g. \cite{Crichton-etal:2020}.

The bioinformatics community recognized that machine learning Relation Extraction (RE) models could help the manual curation task, and the BioCreative workshops introduced the first shared task and manually labeled `gold standard' dataset for training and evaluating models for extracting protein-protein interactions from full text articles in 2006 \cite{Krallinger-etal:2006:BC2PPI}. Several such labeled corpora have followed, primarily focusing on extracting relationships from abstracts. However, labeling of relationships requires domain experts and is slow and expensive. Consequently, most labeled corpora are small, and focus on a small number of entity types and relationships.

In this paper, we introduce {\em ChemDisGene}\footnote{\url{https://github.com/chanzuckerberg/ChemDisGene}}, 
a new dataset of biomedical research abstracts labeled with pairwise interactions between Chemicals, Diseases and Genes/Gene-products. It contains two sub-corpora:
\begin{itemize}[nosep]
\item A large corpus of $\sim80k$ abstracts with distant labeling of 14 relation types. This corpus is automatically derived from CTD \cite{Davis-etal:2020:CTD}, thus allowing for a larger size more suitable for training deep learning models. However, relationships are distantly labeled because relationships in CTD are associated only with a paper, and not with a specific text passage within the paper.

\item A smaller corpus of 523 abstracts, manually annotated with relationships by domain experts. This corpus is aimed primarily for testing models trained on the CTD-derived corpus, and the relationships here are also distantly labeled.
\end{itemize}

A previous version of the CTD-derived corpus was introduced in \cite{Verga-etal:2018:NAACL}. {\em ChemDisGene} adds a manually annotated component, and includes several improvements to the derivation process:
\begin{itemize}[nosep]
\item More recent updates (2021 February) from CTD.

\item Entity linking uses PubTator Central \cite{Wei-etal:2019:PubTator-Central} with significantly improved models for recognizing Chemicals ($+66.3$\% improvement in F1 score), Diseases ($+3.8$\%) and Genes/Proteins ($+8.2$\%) over the previous PubTator model.

\item The previous dataset was randomly split into training, dev and test, while in {\em ChemDisGene} these splits are based on paper publication date, to better simulate a real world scenario.

\item A cleaner extraction of binary relationships from complex nested relationships captured by CTD.
\end{itemize}

The rest of this paper describes how the labeled corpus was developed (\S\ref{sec:methodology}), corpus statistics (\S\ref{sec:corpus_satistics}), baseline models trained and evaluated on the {\em ChemDisGene} (\S\ref{sec:experiments}), and related work (\S\ref{sec:related_work}).

%% file: 03_methodology.tex
\section{Methodology}
\label{sec:methodology}

A note on terminology: we will use {\em relation} to refer to the predicate schema $r(T_s, T_o)$, where $r$ is the {\em relation type}, and $T_s, T_o$ are the argument entity types: {\em Chemical}, {\em Disease} or {\em Gene}. A {\em relationship} is a ground instance $r(e_s, e_o)$ of a {\em relation}, with argument entities $e_s \in T_s, e_o \in T_o$.

The {\em ChemDisGene} dataset comprises a large corpus automatically derived from CTD, and a smaller curated corpus manually labeled by domain experts.

\subsection{Derivation from CTD}
\label{sec:CTD-derivation}

Comparative Toxicogenomics Database (CTD) is a public knowledge base containing manually curated interactions between chemicals, genes, diseases and phenotypes \cite{Davis-etal:2020:CTD}. CTD curators regularly scan new research publications to identify those interactions that are the primary contributions of each paper \cite{Davis-etal:2011:CTD-Curation}. These are then encoded using a hierarchical ontology of $\sim50$ Chemical–Gene interaction classes, and two types each for Chemical–Disease and Gene–Disease interactions (phenotypes are not covered in our dataset). Each interaction is expressed using relation types from these interaction classes, along with the argument entities, and recorded with a reference to the paper from which it was extracted (but no reference to any text within the paper). Entities are also identified using public ontologies: MeSH for Chemicals, MeSH and OMIM for Diseases, and NCBI Gene for Genes and Gene-products\footnote{Links: \href{https://www.nlm.nih.gov/mesh/meshhome.html}{MeSH}, \href{https://www.omim.org}{OMIM}, \href{https://www.ncbi.nlm.nih.gov/gene}{NCBI Gene}}.

While CTD curators scan full papers to extract these relationships, we limited the text in {\em ChemDisGene} to only the title and abstract. Starting with the February 2021 dump of CTD, we obtained abstracts for all referenced articles from PubMed\footnote{\url{https://pubmed.ncbi.nlm.nih.gov}}. Each abstract was processed through PubTator Central\footnote{\url{https://www.ncbi.nlm.nih.gov/research/pubtator/}} (PTC) to identify and link mentions of chemical, disease and gene/gene-product entities. We then performed a `distant alignment' of the annotated abstracts with the relationships linked to each paper in CTD: relationships whose entities were not detected in the abstract were discarded. This yielded a dataset of abstracts with linked entity mentions, and distantly linked relationships.

This distant linking of relationships to aligned abstracts is noisy due to the following sources of error: (i) entity recognition models in PTC, whose F1 scores for each entity type are in the range 0.84--0.90 \cite{Wei-etal:2019:PubTator-Central}, (ii) even if the entities of a relationship are correctly identified in the abstract, the corresponding interaction may not have been mentioned in the abstract text, and (iii) an abstract may mention some relationships that are not extracted by CTD. To measure these sources of error, we selected a subset of aligned abstracts for manual curation (see \S\ref{sec:Curation}, \S\ref{sec:curation-stats}).

Relations in CTD are organized into a class hierarchy, with some relation classes qualified by a `degree'. {\em ChemDisGene} includes 10 of these classes\footnote{Definitions are from the \href{http://ctdbase.org/help/glossary.jsp}{CTD glossary}.}, which combined with the degrees defines 18 relation types:
\begin{itemize}[nosep,leftmargin=10pt]
\itemi{Chemical-Disease}:
\begin{itemize}[nosep,leftmargin=12pt]
\itemi{marker/mechanism}: A chemical that correlates with a disease.
\itemi{therapeutic}: A chemical that has a known or potential therapeutic role in a disease.
\end{itemize}

\itemi{Chemical-Gene}: Each qualified by a {\em degree}.
\begin{itemize}[nosep,leftmargin=12pt]
\itemi{activity}: An elemental function of a molecule. Degrees: {\em increases}, {\em decreases}, or {\em affects} when the direction is not indicated.
%\itemi{binding}: A molecular interaction. Degrees: {\em affects}.
\itemi{binding}: A molecular interaction ({\em affects}).
%\itemi{expression}: Expression of a gene product. Degrees: {\em increases}, {\em decreases}, {\em affects}.
\itemi{expression}: Expression of a gene product ({\em increases}, {\em decreases}, {\em affects}).
%\itemi{localization}: Part of the cell where a molecule resides. Degrees: {\em affects}.
\itemi{localization}: Part of the cell where a molecule resides ({\em affects}).
%\itemi{metabolic\_processing}: The biochemical alteration of a molecule's structure (not including changes in expression, stability, folding, localization, splicing, or transport). Degrees: {\em increases}, {\em decreases}, {\em affects}.
\itemi{metabolic\_processing}: The biochemical alteration of a molecule's structure, not including changes in expression, stability, folding, localization, splicing, or transport ({\em increases}, {\em decreases}, {\em affects}).
%\itemi{transport}: The movement of a molecule into or out of a cell. Degrees: {\em increases}, {\em decreases}, {\em affects}.
\itemi{transport}: The movement of a molecule into or out of a cell ({\em increases}, {\em decreases}, {\em affects}).
\end{itemize}

\itemi{Gene-Disease}:
\begin{itemize}[nosep,leftmargin=12pt]
\itemi{marker/mechanism}: A gene that may be a biomarker of a disease or play a role in the etiology of a disease.
\itemi{therapeutic}: A gene that is or may be a therapeutic target in the treatment of a disease.
\end{itemize}
\end{itemize}
In some cases, CTD defines a finer granularity of Chemical-Gene interactions. Because their occurrence is rare, they would be harder for a model to recognize, so we abstracted them to the levels described above.

The relationships in CTD also include complex and nested biomedical interactions involving multiple entities. For {\em ChemDisGene} we only extracted binary relationships. In particular, (a) we omitted CTD's ``{\em cotreatment}'' relation type because it is non-binary, and (b) we implemented a cleaner extraction of binary relationships from nested interactions (see example in fig.~\ref{fig:binarize}).

The previous CTD-derived dataset in \cite{Verga-etal:2018:NAACL} used the same relation types for Chemical-Disease and Gene-Disease interactions, but a different set of 10 relation types for Chemical-Gene.
%{\em activity-}\{{\em decreases, increases}\}, {\em expression-}\{{\em decreases, increases}\}, {\em metabolic\_processing-}\{{\em decreases, increases}\}, {\em reaction-}\{{\em decreases, increases}\}, {\em response\_to\_substance-affects} and {\em transport-affects}.
With three years of new publications, the distribution of relation types in CTD has changed, affecting our selection.

The derivation of relationships from CTD in \cite{Verga-etal:2018:NAACL} did not take into account nesting levels in complex interactions: in the example in fig.~\ref{fig:binarize}, the previous dataset would also extract {\em reaction-decreases} between the chemical `24-hydroxycholesterol' and the gene `ITGB1', even though the corresponding indicator ``{\em inhibits the reaction}'' is at a different nesting level.

\begin{figure}
\hrule \relax
\vspace{0.1cm}

\begingroup
\fontsize{9pt}{12pt}\selectfont
\begin{tabbing}
\hspace{0.4cm}\=\hspace{0.2cm}\=\hspace{0.2cm}\=\hspace{0.2cm}\=\kill
\textbf{Complex nested interaction in CTD:}\\
\textcolor{Disease}{Quercetin}\tsub{Disease} \emph{inhibits the reaction} \\
\>[ \>\>[ \textcolor{Chemical}{24-hydroxycholesterol}\tsub{Chemical} \emph{co-treated with}\\
\>\>\>\> \textcolor{Chemical}{27-hydroxycholesterol}\tsub{Chemical} \emph{co-treated with}\\
\>\>\>\> \textcolor{Chemical}{cholest-5-en-3 beta,7 alpha-diol}\tsub{Chemical} ]\\
\>\> \emph{results in increased expression of} \textcolor{Gene}{ITGB1}\tsub{Gene} mRNA ]
\end{tabbing}
\endgroup

\vspace{-0.2cm}
\hrule \relax
\vspace{-0.2cm}

\begingroup
\fontsize{9pt}{12pt}\selectfont
\begin{tabbing}
\hspace{0.4cm}\=\kill
\textbf{Extracted binary relationships:}\\
\>{\em expression-increases}(\textcolor{Chemical}{24-hydroxycholesterol}, \textcolor{Gene}{ITGB1})\\
\>{\em expression-increases}(\textcolor{Chemical}{27-hydroxycholesterol}, \textcolor{Gene}{ITGB1})\\
\>{\em expression-increases}(\=\textcolor{Chemical}{cholest-5-en-3 beta,7 alpha-diol},\\
	\>\>\textcolor{Gene}{ITGB1})
\end{tabbing}
\endgroup

\vspace{-0.2cm}
\hrule \relax
\caption{An example showing extraction of binary relationships from a complex nested interaction in CTD.}
\label{fig:binarize}
\end{figure}

As a final step, we added some randomly sampled abstracts that did not align with any CTD relationships as `null' documents with no relationships. This forms 10\% of the CTD-derived corpus, which was then split into \textit{train}, \textit{development} ({\em dev}) and \textit{test} sets by publication year (2018 as dev and years 2019, 2020 as test).

\subsection{Curation}
\label{sec:Curation}

As described above, the relationship labels in the CTD-derived corpus are noisy.
To perform more reliable testing of RE models, we selected some documents for manual annotation: 303 sampled from the {\em test} split, and an additional 252 documents from CTD that were also included in the DrugProt corpus \citelanguageresource{DrugProt}, to enable future comparative analyses. These were distributed for annotation by five biologists, each document assigned randomly to three curators.

We developed a web-based annotation tool which displayed for each document, the title and abstract, all the linked chemicals, diseases and genes/gene-products, and their mentions in the text, and all the relationships derived from CTD. Annotating a document involved two tasks: (i) review each relationship derived from CTD, and either reject or approve it, and (ii) add all other established relationships expressed in the document. Relationships mentioned in the abstract without any conclusions were excluded from annotation. In keeping with our goal of a realistic dataset, 44 of these documents had no CTD-derived relationships.

We developed annotation guidelines (published with the dataset) that describe the steps in the curation process and the types of pairwise interactions curated in this dataset, including brief definitions and real-world example statements that do or do not support a specific relation type. These guidelines underwent multiple rounds of revisions through 4 iterations of practice annotations. During the practice phase, all 5 curators were given the same set of documents to curate (15--30 per cycle); annotation disagreements and questions were clarified during multiple workshops, and feedback and suggestions from the curators were used to improve the guidelines. See fig.~\ref{fig:curation-examples} for some examples.

Some interactions were easy to identify, like {\em Chemical-disease: marker/mechanism}, and were labeled with high consistency. Other relation types required more interpretation and created more disagreement; e.g., the upregulation of a gene product by a chemical can be described by the  types “{\em expression}” or “{\em activity}”, depending on the context. A number of edge cases were identified during the practice phase and added to the guidelines, such as how to record opposite effects, e.g. when both an increase and a decrease in expression of a gene product is mentioned under different experimental conditions, or how to label relationships between two entities that depend on the presence of a third entity.

Only entities correctly detected by PubTator Central and linked to the right ontology record were considered in an interaction pair; the annotation guidelines therefore also included instructions for accepting or rejecting detected entities that did not unambiguously match the text mention, such as a detected entity that is broader than the mention in the abstract.

\begin{figure}
\hrule \relax
\begin{enumerate}[topsep=0pt,itemsep=6pt,label=\textbf{\Alph*:}]
\item {\em Don't record investigated or motivating relationships that remain unknown and hypothetical.}

``\textcolor{Gene}{Gene A} is a therapeutic target for treatment of \textcolor{Disease}{Disease X}; it may therefore have a potential role in treatment of \textcolor{Disease}{Disease Z}.''

{\em Record a relationship between \textcolor{Gene}{Gene A} and \textcolor{Disease}{Disease X}; but not between \textcolor{Gene}{Gene A} and \textcolor{Disease}{Disease Z}.}

\item {\em Inferring a relationship across sentences.}

``We have previously identified a panel of fusion genes in aggressive \textcolor{Disease}{prostate cancers}. In this study, we showed that \ldots \textcolor{Gene}{CCNH}-\textcolor{Gene}{C5orf30} and \textcolor{Gene}{TRMT11}-\textcolor{Gene}{GRIK2} gene fusions were found in \textcolor{Disease}{breast cancer}, \textcolor{Disease}{colon cancer}, \ldots''

{\em Record a `{\em Gene-Disease: marker/mechanism}' relationship between \textcolor{Gene}{C5orf30} and \textcolor{Disease}{prostate cancers}.}
\end{enumerate}
\hrule \relax
\caption{Two examples from the curation guidelines. Colors identify \textcolor{Disease}{disease} and \textcolor{Gene}{gene} mentions.}
\label{fig:curation-examples}
\end{figure}

At the end of the annotation period, we observed that 30 of the documents had been annotated by only two biologists. We also observed that a number of new relationships added to each document had only been added by one annotator. This was not unexpected, as scanning through text to identify all relationships is much harder than verifying whether a specified relationship occurs. We refer to these relationships as `{\em singletons}', and marked as `{\em approved}' all relationships that a majority of the annotators had approved. From the documents annotated by two curators, we also added all relationships derived from CTD that were approved by only one of the two annotators to the list of singletons. We then discarded  documents with more than 10 singletons, while keeping all 252 DrugProt documents, yielding a total of 523 annotated documents.

On analyzing the singletons, we noticed that some of these differed only in degree from an approved relationship in the same document: 45 were abstractions (degree {\em affects}) and 7 refinements (degree {\em increases} or {\em decreases}) of an approved relationship. These singletons were then automatically rejected.

\begin{comment}
In this annotation task, when one annotator does not identify a particular relationship that was found by another, it could be for one of two reasons: (i) both annotators noticed the same text passage but disagreed on whether it expressed the relationship, or (ii) the first annotator did not notice the passage that the second annotator used to identify the relationship. To resolve this ambiguity, we have started a second phase of annotation where all the singleton relationships are reviewed by an annotator not originally assigned to that document. This review phase is in progress, and its results will be published when completed.
\end{comment}
In this annotation task, when one annotator does not identify a particular relationship that was found by another, it could be for one of two reasons: (i) both annotators noticed the same text passage but disagreed on whether it expressed the relationship, or (ii) the first annotator did not notice the passage that the second annotator used to identify the relationship. To resolve this ambiguity, all the singleton relationships were reviewed by an annotator not originally assigned to that document, followed by a second review by the curation manager to ensure consistency. Relationships approved in this phase were added to the curated data.

%% file: 04_corpus.tex
\section{{\em ChemDisGene} Corpus Statistics}
\label{sec:corpus_satistics}

\subsection{The CTD-derived Corpus}

\begin{figure}[!ht]
\begin{center}
\includegraphics[height=5.65cm]{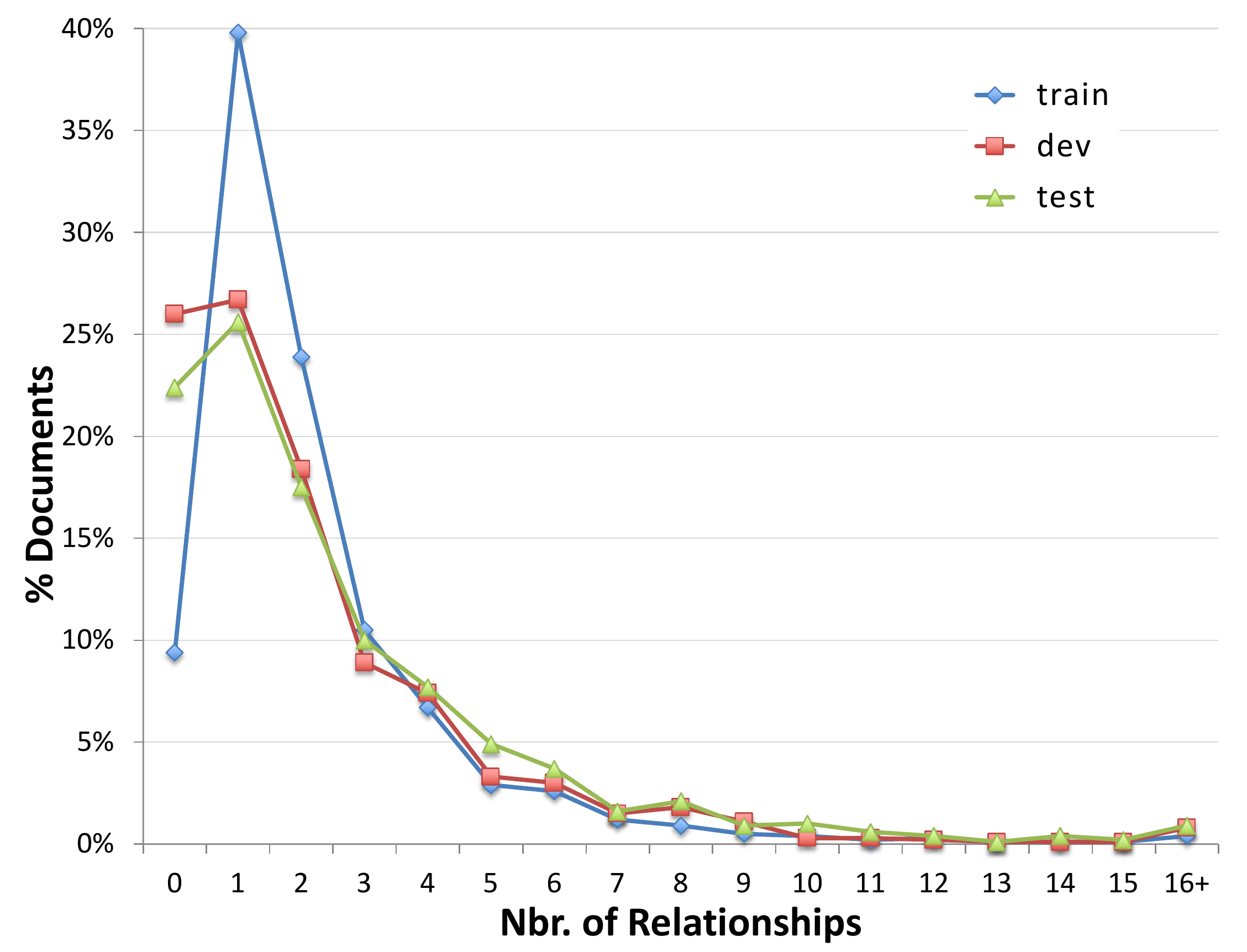}
\caption{Relationships per doc., CTD-derived corpus.}
\label{fig:n_ctd_derived_relns_by_n_docs}
\end{center}
\end{figure}

General statistics for the CTD-derived corpus are shown in table~\ref{tab:CorpusStats}, and the distribution of the number of relationships per document in fig.~\ref{fig:n_ctd_derived_relns_by_n_docs}. About 80\% of the documents have 3 or fewer relationships, followed by a long thin tail. The {\em dev} and {\em test} splits have a higher proportion of documents with no relationships. There are an average of $2.2$ relationships per document, with over 9,000 entity pair occurrences with multiple relation type labels in the same document.

\begin{table}
\begin{center}
\small
\begin{tabular}{lrrr}
\toprule
 				               & \textbf{Train} & \textbf{Dev} & \textbf{Test} \\
\midrule
Nbr. abstracts                      &    76,942 &     1,521 &     1,939 \\
\ldots with no relationships        &     7,244 &       397 &       436 \\
Nbr. relationships                  &   167,005 &     3,290 &     5,116 \\
\ldots unique relationships         &    93,801 &     3,127 &     4,801 \\
Total Entity mentions               & 1,532,117 &    36,114 &    49,839 \\
\hspace{2em} Chemicals              &   686,102 &    13,986 &    19,895 \\
\hspace{2em} Diseases               &   478,397 &     8,962 &    11,750 \\
\hspace{2em} Genes                  &   367,618 &    13,166 &    18,194 \\
Unique Entities in relns.           &    14,991 &     1,894 &     2,345 \\
\hspace{2em} Chemicals              &     7,187 &       759 &       999 \\
\hspace{2em} Diseases               &     2,413 &       283 &       287 \\
\hspace{2em} Genes                  &     5,391 &       852 &     1,059 \\
\bottomrule
\end{tabular}
\end{center}
\caption{\label{tab:CorpusStats}General statistics for the CTD-derived corpus.}
\end{table}

\begin{table*}
\begin{center}
\small
\begin{tabular}{llrrrrrr}
\toprule
		   &                    & \multicolumn{3}{c}{\textit{Total}}  & \multicolumn{3}{c}{\textit{Unique}}\\
\cmidrule(r){3-5} \cmidrule(l){6-8}
\textbf{\#} & \textbf{Relation type} & \textbf{Train} & \textbf{Dev} & \textbf{Test} & \textbf{Train} & \textbf{Dev} & \textbf{Test} \\
\midrule
 1 & Chemical-Disease : marker/mechanism                & 66,155 &   559 &   754 & 27,706 &   486 &   602 \\
 2 & Chemical-Disease : therapeutic                     & 34,775 &   250 &   410 & 16,093 &   245 &   398 \\
 3 & Chemical-Gene : activity - decreases               &  5,555 &   101 &   232 &  4,128 &    97 &   232 \\
 4 & Chemical-Gene : activity - increases               &  6,152 &   127 &   174 &  4,133 &   120 &   157 \\
 5 & Chemical-Gene : binding - affects                  &  3,123 &    67 &    77 &  2,024 &    65 &    73 \\
 6 & Chemical-Gene : expression - affects               &  1,247 &    51 &   160 &  1,206 &    51 &   158 \\
 7 & Chemical-Gene : expression - decreases             & 10,204 &   480 &   923 &  8,487 &   467 &   905 \\
 8 & Chemical-Gene : expression - increases             & 19,810 &   919 & 1,570 & 14,685 &   878 & 1,491 \\
 9 & Chemical-Gene : localization - affects             &  1,448 &    50 &    73 &  1,216 &    50 &    70 \\
10 & Chemical-Gene : metabolic\_processing - decreases  &  1,653 &   101 &   116 &  1,313 &   100 &   111 \\
11 & Chemical-Gene : metabolic\_processing - increases  &  4,640 &   175 &   293 &  3,507 &   172 &   283 \\
12 & Chemical-Gene : transport - increases              &  1,962 &    92 &   108 &  1,405 &    88 &    96 \\
13 & Gene-Disease : marker/mechanism                    &  9,388 &   301 &   219 &  7,384 &   292 &   218 \\
14 & Gene-Disease : therapeutic                         &    893 &    17 &     7 &    514 &    16 &     7 \\
\bottomrule
\end{tabular}
\end{center}
\caption{\label{tab:RelnFreq-CTD-derived}Nbr. of relationships (instances) for each relation type in the CTD-derived corpus.}
\end{table*}

Counts for each relation type are shown in table~\ref{tab:RelnFreq-CTD-derived}. Unique numbers count unique argument-entity pairs. Four Chemical-Gene relation types ({\em activity-affects}, {\em metabolic\_processing-affects}, {\em transport-affects}, and {\em transport-increases}) were omitted from the CTD-derived corpus because of their low incidence. However they are included in the curated corpus for completeness, making the annotation task a little easier.

\subsection{The Curated Corpus}
\label{sec:curation-stats}

\begin{table*}
\begin{center}
\small
\begin{tabular}{llrr}
\toprule
		   &                    & \multicolumn{2}{c}{\textit{Distribution} (\%)} \\
\cmidrule{3-4}
\textbf{\#} & \textbf{Relation type} & \textbf{Approved, New} & \textbf{Approved, CTD} \\
\midrule
 1 & Chemical-Disease : marker/mechanism               &     16.6 &     16.4 \\
 2 & Chemical-Disease : therapeutic                    &     10.4 &     12.0 \\
 3 & Chemical-Gene : activity - affects                &      1.2 &          \\
 4 & Chemical-Gene : activity - decreases              &      8.3 &      7.3 \\
 5 & Chemical-Gene : activity - increases              &      8.7 &      7.8 \\
 6 & Chemical-Gene : binding - affects                 &      4.3 &      6.7 \\
 7 & Chemical-Gene : expression - affects              &      2.8 &      0.6 \\
 8 & Chemical-Gene : expression - decreases            &     10.4 &     13.1 \\
 9 & Chemical-Gene : expression - increases            &     11.8 &     18.4 \\
10 & Chemical-Gene : localization - affects            &      0.8 &      1.5 \\
11 & Chemical-Gene : metabolic\_processing - affects   &      0.8 &          \\
12 & Chemical-Gene : metabolic\_processing - decreases &      1.7 &      1.5 \\
13 & Chemical-Gene : metabolic\_processing - increases &      3.0 &      4.0 \\
14 & Chemical-Gene : transport - affects               &      0.3 &          \\
15 & Chemical-Gene : transport - decreases             &      0.6 &          \\
16 & Chemical-Gene : transport - increases             &      1.1 &      0.9 \\
17 & Gene-Disease : marker/mechanism                   &     14.1 &      9.3 \\
18 & Gene-Disease : therapeutic                        &      2.9 &      0.5 \\
\bottomrule
\end{tabular}
\end{center}
\caption{\label{tab:RelnFreq-Curated}Frequency distribution of relation types in curated corpus (each column sums to 100\%). Empty frequencies indicate some relations are rare in CTD.}
\end{table*}

The curated corpus contains 523 documents: 271 from CTD-derived's {\em test} split, and an additional 252 documents taken from DrugProt, that are not in the CTD-derived corpus. Twenty seven of these documents had no relationships derived from CTD. Manual annotation rejected 22\% of all CTD-derived relationships, leaving 64 documents with no approved CTD-derived relationships. This indicates a fairly high 78\% confidence in the automatically derived relationships.

There are a total of 1,279 approved CTD-derived relationships (avg. 2.4/doc), 2,632 approved new relationships (5.0/doc). The distribution of the 18 types of relations in this corpus is shown in table~\ref{tab:RelnFreq-Curated}.

\begin{figure}
\begin{center}
\includegraphics[height=5.65cm]{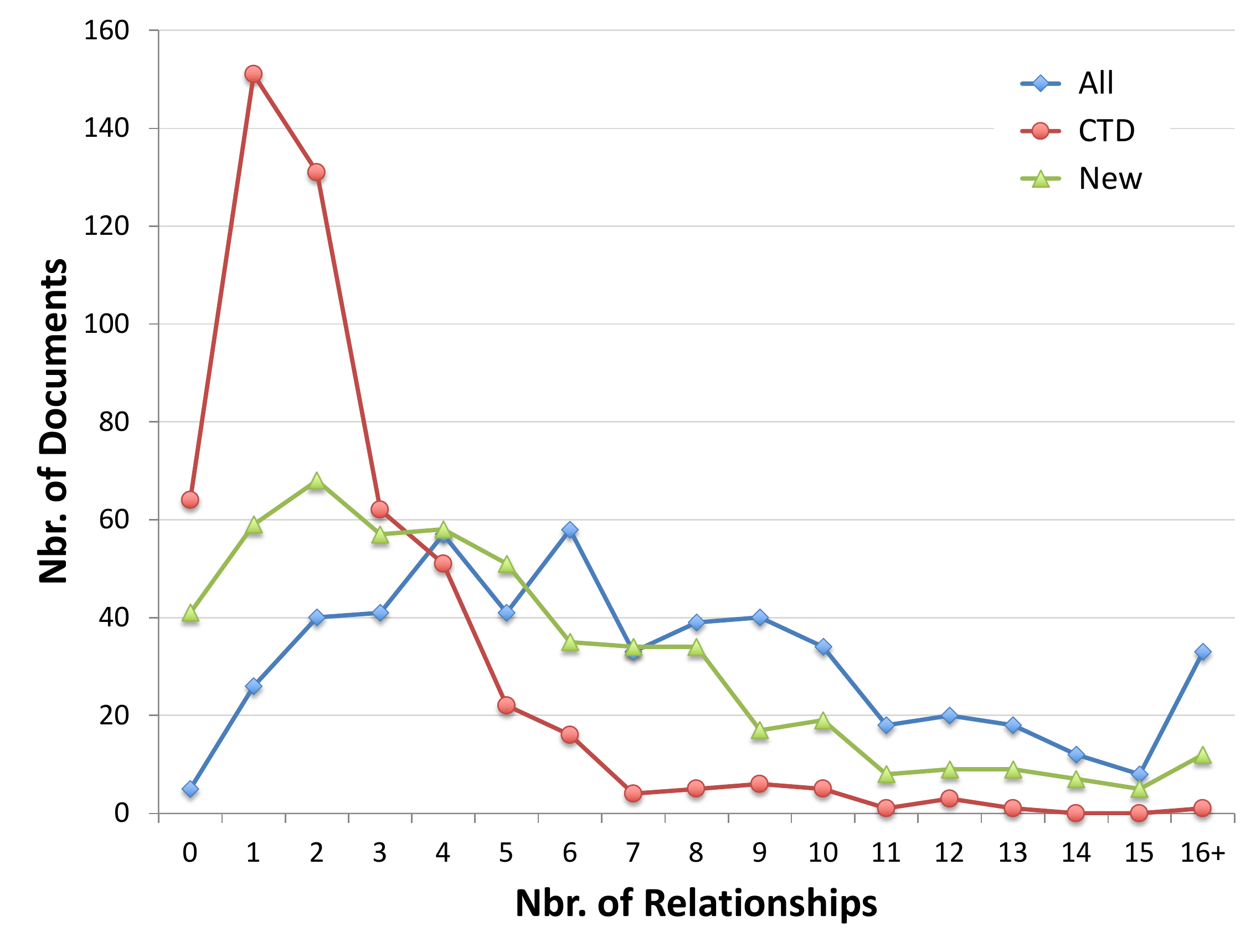}
\caption{Distribution of the nbr. of approved relationships per document in the Curated corpus.}
\label{fig:n_apprvd_relns_by_n_docs}
\end{center}
\end{figure}

The 3,911 approved relationships (3,806 unique) in the curated corpus involve 1,875 unique entities: 670 unique Chemicals, 318 Diseases and 887 Genes. Figure~\ref{fig:n_apprvd_relns_by_n_docs} shows the distribution of number of approved relationships in each document. As expected, the CTD-derived approved relationships are more skewed to the left than the new added relationships.

\subsection{Inter-Annotator Agreement}

Commonly used measures of inter-annotator agreement are defined for tasks where the units being classified or measured are precisely specified. As noted in \cite{Kilicoglu-etal:2011}, identifying all relationships expressed in a text does not match this paradigm. This task could be decomposed into the following steps: (i) find relationship indicators in the text, (ii) identify the entity mentions each indicator refers to, and (iii) map the expressed relationship to the appropriate ontological term. Here the space of possible annotations is clearly defined only for step (iii). In step (ii) the space would be clearly specified only if we presented the annotators with every pair of linked mentions. The set of possible relationship indicators in a document, in step (i), is also not presented to the annotators.
When a relationship is identified by only one of two curators reviewing the same text, it could be because either the first one did not `notice' the same sentence, or actually saw it and rejected it.
This inherent ambiguity causes a problem even for measures that allow varying number of annotations per unit.

Similar to \cite{Kilicoglu-etal:2011}, we evaluated each curator's annotations against a reference, using precision, recall and F1 scores, as more feasible and intuitively understandable metrics for our use case. We used the `majority approved' relationships (\S\ref{sec:Curation}) as the reference dataset. The annotator agreement metrics (table~\ref{tab:AnnotrAgrmt}) are fairly high, indicating a high confidence in the approved subset. As expected, agreement levels on prompted relationships (those from CTD) in the annotator UI is higher than for relationships that the annotator has to find and add (new relationships).

%\begin{table}
%\begin{center}
%\small
%\begin{tabular}{lrrr}
%\toprule
%\textbf{Annotator} & \textbf{Precision} & \textbf{Recall} & \textbf{F1} \\
%\midrule
%\multicolumn{4}{l}{{\em \hspace{2em}All relationships}:} \\
%A & 0.849 & 0.856 & 0.852 \\
%B & 0.830 & 0.852 & 0.841 \\
%C & 0.853 & 0.813 & 0.832 \\
%D & 0.825 & 0.944 & 0.881 \\
%E & 0.876 & 0.890 & 0.883 \\
%\midrule
%\multicolumn{4}{l}{{\em \hspace{2em}Relationships from CTD only}:} \\
%A & 0.999 & 0.980 & 0.989 \\
%B & 0.999 & 0.920 & 0.958 \\
%C & 1.000 & 0.941 & 0.970 \\
%D & 1.000 & 0.984 & 0.992 \\
%E & 1.000 & 0.921 & 0.959 \\
%\midrule
%\multicolumn{4}{l}{{\em \hspace{2em}New relationships only}:} \\
%A & 0.753 & 0.773 & 0.763 \\
%B & 0.729 & 0.804 & 0.765 \\
%C & 0.705 & 0.681 & 0.693 \\
%D & 0.740 & 0.919 & 0.820 \\
%E & 0.800 & 0.868 & 0.833 \\
%\bottomrule
%\end{tabular}
%\end{center}
%\caption{\label{tab:AnnotrAgrmt}Agreement metrics for each annotator against the `approved' reference subset. \sunil{We can compress this to F1 scores only to save space, if needed.}}
%\end{table}

\begin{table}
\begin{center}
\small
\begin{tabular}{lrrrrr}
\toprule
Relationships    & \textbf{A} & \textbf{B} & \textbf{C} & \textbf{D} & \textbf{E} \\
\midrule
All              &       0.85 &       0.84 &       0.83 &       0.88 &       0.88 \\
CTD-derived only &       0.99 &       0.96 &       0.97 &       0.99 &       0.96 \\
New only         &       0.76 &       0.77 &       0.69 &       0.82 &       0.83 \\
\bottomrule
\end{tabular}
\end{center}
\caption{\label{tab:AnnotrAgrmt}Agreement F1 scores for the 5 annotators (A-E) against the `approved' reference subset.}
\end{table}

%% file: 05_re_task.tex
\section{The Relationship Extraction Task}
\label{sec:experiments}

\subsection{Task definition}

The document-level relation extraction (RE) task in {\em ChemDisGene} is to identify all relationships $r(e_s, e_o)$ expressed in a document, comprised of the title and abstract texts, that are the primary contributions of that article. We consider 14 binary relation types (from the CTD-derived corpus) among chemical, disease and gene/gene-product entities. All mentions of these entities in the text are identified and linked to the corresponding ontologies. This is a {\em distant supervision} (relationships are associated with documents, but not specific entity mentions) {\em multi-label} (a document, and a pair of entities, may have more than one relationship) classification task. For evaluation, we use Micro/Macro F1 scores where per-relation thresholds are tuned on the dev set, and average precision where thresholds are not required.

\subsection{Models}

In our experiments, we trained and evaluated three baseline methods on \emph{ChemDisGene}.

\noindent \textbf{BRAN} %~\footnote{We do not train the NER joint loss for BRAN model.}
Bi-affine relation attention networks \cite{Verga-etal:2018:NAACL}
is one of the first papers to tackle document level (distant supervision) relation extraction in the biomedical domain. It uses multiple self-attention + convolutional neural network (NN) layers to encode the text input, then leverages per-relation biaffine transformation to calculate mention level scores of the query $r(e_s, e_o)$, and a $logsumexp$ layer to capture the most significant signal among mention pairs. In our experiment, we omitted BRAN's NER joint loss in order to analyze its core RE module.
%The model structure looks like: $Input \rightarrow transformer \rightarrow CNNs \rightarrow transformer \rightarrow CNNs \rightarrow Biaffine \rightarrow logsumexp \rightarrow logits $

\noindent \textbf{PubmedBert}~\cite{gu2021domain} is a BERT-based pretrained language model \cite{Devlin-etal:2019:BERT-ACL} trained from scratch on PubMed abstracts.
For relation extraction, we first get each entity's embeddings by max-pooling over PubmedBert's encoding of all the entity's mentions. Then concatenated embeddings of candidate argument entity pairs are processed through a feed-forward NN to predict scores for each relation type.
%The model structure looks like: $Input \rightarrow BertEncoder \rightarrow FFNs \rightarrow logits$

\noindent \textbf{PubmedBert + BRAN}. This model combines the stronger text encoder of PubmedBert with the relation detection layers of BRAN.
%In order to leverage the benefit of pretrained language model, and also be able to select the significant entity mention pairs from the abstract, we combined the above baselines into a stronger model.
The model structure is: \textit{Input} $\rightarrow$ \textit{PubMedBertEncoder} $\rightarrow$ \textit{Biaffine} $\rightarrow$ \textit{logsumexp} $\rightarrow$ \textit{logits}.

\subsection{Empirical Results}

Table~\ref{tab:PerformanceModels} shows overall performance of the three baseline models on \emph{ChemDisGene}\footnote{We trained all three baselines on $ChemDisGene$ training set with hidden dimension 128, and we tuned the hyper-parameters such as learning rate [1e-5, 1e-4] and weight decay = [0, 1e-4] over the distant supervision dev set.}. Performance metrics are shown for the test split of the CTD-derived corpus, and separate metrics on the curated corpus for approved relationships derived from CTD, and for all approved relationships, which also includes new relationships added by the curators.

From the table, our best model PubmedBert + BRAN has 43.8 Micro F1 and 50.6 average precision on the `\emph{all relationships}' curated test set, indicating the difficulty of this task. The pretrained language model adds significant improvement over BRAN. And the biaffine transformation and $logsumexp$ layer are also complementary to the pretrained language model.

% While PubmedBert + BRAN out-performs PubmedBert on both F1 and Average Precision on CTD-derived test set, its F1 value drops on the \emph{CTD only} curated test set. This suggests that thresholds tuned on the derived dev set are not well suited for the manually annotated ground truth labels.

Compared with the CTD-derived test set, the performance decreases significantly on the curated test set, indicating the necessity of evaluation on expert-labeled data.
We also observe that Macro results are lower than Micro, indicating that performance varies across different relation types.
In table~\ref{tab:PerformanceByReln} we see that relation types with low frequencies in the training data tend to perform poorly. The particularly bad performance of our model on {\em Chemical-Gene: expression-affects} is also caused by distraction from two similar but common {\em Chemical-Gene} relation types: {\em expression-increases} and {\em expression-decreases}.

\paragraph{Performance of baseline models on `BRAN' dataset.} We also trained and tested the baseline models on the CTD-derived dataset from \cite{Verga-etal:2018:NAACL}, referred to as the `BRAN' dataset (table~\ref{tab:PerfBranCTD}). As described above (\S\ref{sec:CTD-derivation}), there are several differences in this dataset that account for the different performance results from those on {\em ChemDisGene} (table~\ref{tab:PerformanceModels}). Perhaps the most important one is that in the BRAN dataset, abstracts from {\em test} and {\em dev} splits are randomly selected, whereas in {\em ChemDisGene} 271 abstracts are assigned based on publication date. The {\em ChemDisGene} test set also includes more documents with no relationships. While this makes {\em ChemDisGene} more challenging, it also reflects a more realistic scenario for applying such RE models. The relative order of performance of the three baselines is the same on both datasets.

\paragraph{Comparing the performance on CTD-derived and All relationships in curated corpora.}
From Table~\ref{tab:PerformanceModels} we can see that while model precision increases when tested on {\em all approved relationships} from the curated corpus, compared to the performance on just {\em CTD-derived approved relationships}, the recall of all models drops significantly. A main reason is that the training data only includes CTD-derived relationships, which are selected by CTD to be the `primary' contributions of the paper. While this is mostly determined within the context of other publications, there might be a signal in the wording (an area for further investigation).

Curators were asked to reject CTD-derived relationships when the entities involved were incorrectly linked. This probably accounts for the small difference in models' performance between the CTD-derived and curated corpora.

\begin{table*}
\begin{center}
\small
\begin{tabular}{l|rrrr|rrr}
\toprule
 & \multicolumn{4} {c|} {Micro} & \multicolumn{3} {c} {Macro} \\
\textbf{Model} & \textbf{P} & \textbf{R} & \textbf{F1} & \textbf{Avg. P} & \textbf{P} & \textbf{R} & \textbf{F1}\\

\midrule
\midrule
\multicolumn{8}{c}{{\emph{CTD-derived corpus: `dev' split / `test' split}}} \\
% % BRAN & 32.1 & 46.3 & 37.9 &  28.4 & 25.9 & 32.3 & 28.2 \\
% % PubmedBert &  49.8 & 60.3 & 54.5 & 49.9 & 42.2 & 52.4 & 46.1 \\
% % PubmedBert + BRAN &  53.9 & 61.0 & 57.3 & 54.0 & 45.0 & 54.1 & 48.7 \\
\midrule
BRAN & 32.1 / 31.7 & 46.3 / 44.2 & 37.9 / 36.9 & 28.4 / 27.9 & 25.9 / 23.6 & 32.3 / 30.1 & 28.2 / 26.0 \\
% PubmedBert & 49.8 / 49.8 & 60.3 / 58.7 & 54.5 / 53.9 & 49.9 / 51.3 & 42.2 / 41.5 & 52.4 / 49.7 & 46.1 / 44.3 \\
PubmedBert & 50.3 / 49.6 & 59.3 / 56.1 & 54.5 / 52.6 & 50.3 / 50.1 & 43.6 / 39.0 & 50.3 / 48.4 & 44.9 / 41.7 \\
PubmedBert + BRAN & 53.9 / 53.9 & 61.0 / 57.3 & 57.3 / 55.6 & 54.0 / 54.3 & 45.0 / 42.7 & 54.1 / 50.4 & 48.7 / 44.4 \\
\midrule
\midrule
\multicolumn{8}{c}{{\emph{Curated corpus: CTD-derived relationships only / All relationships}}} \\
\midrule
% BRAN              & 29.6 / 41.2 & 43.4 / 28.0 & 35.2 / 33.4 & 31.7 / 33.1 & 25.0 / 36.0 & 33.7 / 23.5 & 27.1  / 26.1 \\
% PubmedBert        & 49.3 / 62.9 & 63.6 / 37.2 & 55.5 / 46.7 & 54.8 / 47.6 & 46.2 / 58.2 & 59.6 / 36.6 & 50.9  / 42.4 \\
% PubmedBert + BRAN & 51.5 / 66.3 & 59.7 / 35.1 & 55.3 / 45.9 & 57.5 / 51.5 & 51.1 / 65.2 & 56.3 / 35.7 & 48.8  / 41.1 \\

% BRAN & 27.0 / 41.7 & 43.1 / 28.0 & 33.2 / 33.5 & 28.8 / 33.7 & 22.9 / 35.9 & 33.7 / 23.4 & 25.5 / 25.8 \\
% PubmedBert & 45.0 / 62.7 & 63.6 / 37.3 & 52.7 / 46.8 & 51.5 / 47.6 & 41.0 / 57.5 & 59.6 / 36.8 & 47.0 / 42.1 \\
% PubmedBert + BRAN & 47.4 / 66.5 & 59.5 / 35.2 & 52.7 / 46.0 & 53.0 / 51.4 & 46.5 / 64.6 & 56.6 / 35.7 & 45.8 / 40.8 \\

BRAN & 24.4 / 41.8 & 45.8 / 26.6 & 31.8 / 32.5 & 28.1 / 33.5 & 20.3 / 37.2 & 35.7 / 22.5 & 24.5 / 25.8 \\
% PubmedBert & 52.0 / 72.6 & 40.8 / 19.3 & 45.7 / 30.6 & 46.7 / 45.6 & 44.2 / 67.4 & 30.4 / 15.7 & 34.0 / 24.2 \\
PubmedBert & 43.0 / 64.3 & 61.7 / 31.3 & 50.7 / 42.1 & 50.7 / 46.9 & 34.7 / 53.7 & 53.4 / 32.0 & 39.6 / 37.0 \\
PubmedBert + BRAN & 46.5 / 70.9 & 61.1 / 31.6 & 52.8 / 43.8 & 53.0 / 50.6 & 45.8 / 69.8 & 59.0 / 32.5 & 47.0 / 40.5 \\
\bottomrule
\end{tabular}
\end{center}
\caption{\label{tab:PerformanceModels} Performance of baseline models on {\em ChemDisGene} CTD-derived `test' and curated corpora.}
\end{table*}

\begin{table}
\begin{center}
\small
\begin{tabular}{lr}
\toprule
\textbf{Relation Type}  & \textbf{F1} \\
\midrule
Chemical-Disease : marker/mechanism & 54.1 \\
Chemical-Disease : therapeutic & 45.5 \\
Chemical-Gene : expression - increases & 58.2 \\
Chemical-Gene : expression - decreases & 61.6 \\
Gene-Disease : marker/mechanism & 47.1 \\
Chemical-Gene : activity - increases & 52.4\\
Chemical-Gene : activity - decreases & 56.3 \\
Chemical-Gene : metabolic\_processing - increases & 36.4 \\
Chemical-Gene : binding - affects  & 58.1 \\
Chemical-Gene : transport - increases & 36.1 \\
Chemical-Gene : metabolic\_processing - decreases & 34.4 \\
Chemical-Gene : localization - affects  & 48.9 \\
Chemical-Gene : expression - affects  & 0.4 \\
Gene-Disease : therapeutic & 28.6 \\
\bottomrule
\end{tabular}
\end{center}
%\caption{Performance of the `PubmedBert + BRAN' model on each relation type in the curated corpus. Rows are sorted from up to down with decreasing frequencies on the training data.}
\caption{`PubmedBert + BRAN' model metrics for each relation type in the curated corpus, sorted on decreasing relation frequency in the training data.}
\label{tab:PerformanceByReln}
\end{table}

\begin{table}
\begin{center}
\small
\begin{tabular}{l|rr}
\toprule
%\textbf{Model} & \textbf{Precision} & \textbf{Recall} & \textbf{F1} & \textbf{AP} \\
\textbf{Model} & \textbf{Micro F1} & \textbf{Macro F1}\\
\midrule
BRAN &  43.5 &   30.1\\
PubmedBert &  58.9 & 44.6  \\
PubmedBert + BRAN & 60.0 & 46.0  \\
\bottomrule
\end{tabular}
\end{center}
\caption{\label{tab:PerfBranCTD}Evaluation on data from (Verga et al., 2018).}
% including \cite{Verga-etal:2018:NAACL} in caption causes an error
\end{table}

%% file: 02_related_work.tex
\section{Related Work}
\label{sec:related_work}

\subsection{Distant Supervision Biomedical Corpora}

As described above (see \S\ref{sec:intro},\ref{sec:CTD-derivation}), {\em ChemDisGene} offers a reworking of the derived corpus introduced in \cite{Verga-etal:2018:NAACL}, focusing on a cleaner derivation from an updated CTD with better entity linking. The number of abstracts also increased by $\sim20k$.

A well known manually labeled biomedical corpus is BC5-CDR \cite{Li-etal:2016:BC5CDR}, which identifies a single relation type between Chemicals and Diseases, distantly labeled in 1,500 abstracts. {BC6-PM} \cite{Dogan-etal:2019:BC6-PM} is another manually annotated distant supervision corpus, for Protein-Protein interactions. It has a total of 1,232 abstracts, but only one relation type.

The GDA dataset \cite{Wu-etal:2019:RENET} takes a similar approach to CTD-derived, to derive a Gene-Disease associations dataset from the DisGeNET\footnote{\url{https://www.disgenet.org}} database, using PubTator to link entity mentions. Abstracts are distantly labeled with a single relation type.

\subsection{Direct Supervision Biomedical Corpora}

DrugProt \cite{Miranda-etal:2021:DrugProt}, \citelanguageresource{DrugProt}, is the most recent manually annotated corpus of biomedical research abstracts covering multiple (13) relation types between Chemicals and Genes/Gene-products. {\em ChemDisGene} uses a different set of 14 relation types between Chemicals and Genes, derived from CTD. These relations generally describe the observed effect of an interaction. For example, a {\em Localization} relation is recorded when the interaction between a Chemical and Gene product affects the part of the cell where the molecule resides. In contrast, the DrugProt relation classes are defined by the specific type of interaction between a Chemical and Gene/Gene-product: they distinguish between `Direct' and `Indirect' regulation (where possible), and the subclasses focus on the direction of the interaction (`Upregulator', `Downregulator'). The subclasses for direct regulation are highly granular, differentiating between `Activator', `Agonist', `Antagonist', etc. %`Agonist-Activator' and so on.

As an example, the relationship expressed in the text
\begingroup
\small
\vspace{-1ex}
\addtolength\leftmargini{-0.5cm}
\begin{quote}
``\texttt{\textcolor{Chemical}{bisphenol P}\tsub{Chemical} showed \textcolor{Gene}{estrogen receptor}\tsub{Gene} antagonistic activities}''
\end{quote}
\vspace{-1ex}
\endgroup
would be annotated as {\em Chemical-Gene: activity-decreases} in {\em ChemDisGene}, whereas DrugProt would record it as {\em Chemical-Gene: antagonist}.

DrugProt contains a larger number of abstracts (3500 in training, 750 in dev), with $\sim$5 relationships per abstract. All mentions of Chemicals and Gene-related entities are identified, but not linked. Relationships are {\em directly supervised} by identifying the actual pair of mentions expressing each relationship.

Our {\em ChemDisGene} manually curated corpus is smaller, but also includes relationships between Chemicals and Diseases, and Diseases and Genes. All entity mentions are identified and linked by the models in Pubtator Central, and relationships are distantly labeled, associated with a document but not specific entity mentions. The curated corpus contains $\sim$6 approved relationships per abstract, distinguishing between primary contributions (derived from CTD) and other (`new') relationships.

Most other manually annotated corpora used in biomedical RE tasks are also directly supervised, and cover fewer relation types, typically between fewer types of entities. As another example, Drug-drug interaction (DDI)~\cite{herrero2013ddi} specifies 4 relation types among drugs, on sentences extracted from 1025 documents.

\subsection{Other RE Corpora}

In the general domain, there exist several RE benchmarks for sentence level, document level and few-shot scenarios.
SemEval 2010 task 8~\cite{hendrickx-etal-2010-semeval} includes ten semantic relation types between nouns over $\sim11k$ sentences.
The TAC relation extraction dataset (TACRED)~\cite{zhang2017tacred},  as used in the TAC KBP challenges, contains 106k sentences from newswire and web text covering 41 relation types. TACREV~\cite{alt2020tacred} and Re-TACRED~\cite{stoica2021re} provides cleane versions of TACRED.
DOCRED~\cite{yao-etal-2019-docred} is a document level relation extraction dataset on the WikiPedia domain, with 5053 manually annotated documents and 100 relation types.
FewRel \cite{han2018fewrel}is a relation extraction benchmark for few-shot scenario, based on WikiPedia. A newer version \cite{Gao-etal:2019:FewRel2} includes Biomedical relations as a domain adaptation task.

\subsection{Relation Extraction Models}

Traditional RE models have focused on classifying the entity interaction in a sentence.
For example, ~\newcite{zeng-etal-2014-relation} encoded sentences and entity pairs with convolutional neural networks and position embeddings. ~\newcite{soares2019matching} finetuned Bert with self-supervised signals from entity linking, and applied the model to downstream RE tasks. 
There is also previous work targeting longer text passages such as cross sentence RE~\cite{quirk2017distant}, or document level distant supervision RE~\cite{Verga-etal:2018:NAACL,sahu2019inter,christopoulou2019connecting}.

\newcite{sahu2019inter} and \newcite{christopoulou2019connecting} encode graphs generated from each document for RE. In contrast, BRAN \cite{Verga-etal:2018:NAACL}
%leverages distant supervision signals from CTD and tackles multiple mentions of each entity pair via biaffine transformation and \emph{logsumexp} layer to select the most significant signals. These document level RE model are all tested over BC5CDR dataset with comparable performances. Due to the simplicity of BRAN, which does not rely on extra graph generation step, we choose it as our baseline model.
uses transformers to encode the text sequence and then evaluates each mention pair of candidate argument entities. All these document level RE models showed comparable performance on BC5CDR. We chose BRAN as a baseline because it does not require a graph generation step.

In addition to RE, there is triple extraction work \cite{Bansal-etal:2020} that recognizes entities and relationships simultaneously.

There has also been recent work on extracting $n$-ary biomedical relationships across sentences, e.g. \cite{Ernst-etal:2018} learns dependency parse tree patterns from seed facts, and
\cite{peng2017cross} applies graph LSTMs to dependency parses, trained on a noisy distant labeled dataset. In this paper we focus on binary relations.

%% file: 06_conclusion.tex
\section{Conclusion}

We introduced {\em ChemDisGene}, a new dataset of research abstracts labeled with biomedical entity mentions and distance-labeled with biomedical relationships, for training and evaluating multi-type multi-label biomedical RE models. The dataset includes a large automatically derived corpus with noisy relationship labels ($\sim22$\% noise based on manual curation), and a cleaner manually curated dataset of 523 abstracts. We also provided three baseline ML models for RE, trained and evaluated on the {\em ChemDisGene} dataset.
We believe this is the first dataset for biomedical relation extraction tasks that addresses multiple entity (more than 2) and relation types, 
and includes both a large automatically derived corpus (useful for model training), as well as a smaller corpus labeled by human experts.

Manually annotating raw text with biomedical relationships is a hard and time consuming task, even for domain experts. We facilitated the curation with high quality models for entity recognition. 

Future refinements to this dataset could include verifying the linked entities in the curated corpus, and adding {\em Protein-Protein} interactions, useful for understanding disease mechanisms and drug repurposing.